\documentclass[12pt]{article}
\usepackage{chicagor}
\usepackage{times}

\ifx\pdfoutput\undefined
  \usepackage[dvips]{color}
  \usepackage{graphicx}
  \DeclareGraphicsExtensions{.eps}
\else
 \usepackage[pdftex]{color}
 \usepackage[pdftex]{graphicx}
 \DeclareGraphicsExtensions{.png}
\fi
\newcommand{\heads}{\mathit{heads}}
\newcommand{\tails}{\mathit{tails}}
\newcommand{\Monday}{\mathit{Monday}}

\newtheorem{THEOREM}{Theorem}[section]
\newenvironment{theorem}{\begin{THEOREM} \hspace{-.85em} {\bf :} }%
                        {\end{THEOREM}}
\newtheorem{LEMMA}[THEOREM]{Lemma}
\newenvironment{lemma}{\begin{LEMMA} \hspace{-.85em} {\bf :} }%
                      {\end{LEMMA}}
\newtheorem{COROLLARY}[THEOREM]{Corollary}
\newenvironment{corollary}{\begin{COROLLARY} \hspace{-.85em} {\bf :} }%
                          {\end{COROLLARY}}
\newtheorem{PROPOSITION}[THEOREM]{Proposition}
\newenvironment{proposition}{\begin{PROPOSITION} \hspace{-.85em} {\bf :} }%
                            {\end{PROPOSITION}}
\newtheorem{DEFINITION}[THEOREM]{Definition}
\newenvironment{definition}{\begin{DEFINITION} \hspace{-.85em} {\bf :} \rm}%
                            {\end{DEFINITION}}
\newtheorem{CLAIM}[THEOREM]{Claim}
\newenvironment{claim}{\begin{CLAIM} \hspace{-.85em} {\bf :} \rm}%
                            {\end{CLAIM}}
\newtheorem{EXAMPLE}[THEOREM]{Example}
\newenvironment{example}{\begin{EXAMPLE} \hspace{-.85em} {\bf :} \rm}%
                            {\end{EXAMPLE}}
\newtheorem{REMARK}[THEOREM]{Remark}
\newenvironment{remark}{\begin{REMARK} \hspace{-.85em} {\bf :} \rm}%
                            {\end{REMARK}}

\newcommand{\thm}{\begin{theorem}}
\newcommand{\lem}{\begin{lemma}}
\newcommand{\pro}{\begin{proposition}}
\newcommand{\dfn}{\begin{definition}}
\newcommand{\rem}{\begin{remark}}
\newcommand{\xam}{\begin{example}}
\newcommand{\cor}{\begin{corollary}}

\newcommand{\ethm}{\end{theorem}}
\newcommand{\elem}{\end{lemma}}
\newcommand{\epro}{\end{proposition}}
\newcommand{\edfn}{\bbox\end{definition}}
\newcommand{\erem}{\bbox\end{remark}}
\newcommand{\exam}{\bbox\end{example}}
\newcommand{\ecor}{\end{corollary}}

\newcommand{\beqn}{\begin{equation}}
\newcommand{\eeqn}{\end{equation}}

\newcommand{\bbox}{\vrule height7pt width4pt depth1pt}

\newcommand{\clm}{\begin{claim}}
\newcommand{\eclm}{\end{claim}}

\newcommand{\inter}{\cap}

\renewcommand{\phi}{\varphi}

\newcommand{\K}{{\cal K}}

\newcommand{\R}{{\cal R}}

\newcommand{\<}{\langle}
\renewcommand{\>}{\rangle}

\newcommand{\ol}{\setlength{\itemsep}{0pt}\begin{enumerate}}
\newcommand{\eol}{\end{enumerate}\setlength{\itemsep}{-\parsep}}
\newcommand{\ul}{\setlength{\itemsep}{0pt}\begin{itemize}}
\newcommand{\dl}{\setlength{\itemsep}{0pt}\begin{description}}
\newcommand{\edl}{\end{description}\setlength{\itemsep}{-\parsep}}
\newcommand{\eul}{\end{itemize}\setlength{\itemsep}{-\parsep}}

\newcommand{\commentout}[1]{}

\newcommand{\bi}{\begin{itemize}}
\newcommand{\ei}{\end{itemize}}
\newcommand{\be}{\begin{enumerate}}
\newcommand{\ee}{\end{enumerate}}

\begin{document}
\newcommand{\tick}{\mbox{\it tick}}
\renewcommand{\S}{{\cal S}}

\title{Sleeping Beauty Reconsidered:\\ Conditioning and Reflection in
Asynchronous Systems}
\author{Joseph Y. Halpern%
\thanks{Work supported in part by NSF under grant 
CTC-0208535, by ONR under grants  N00014-00-1-03-41 and
N00014-01-10-511, by the DoD Multidisciplinary University Research
Initiative (MURI) program administered by the ONR under
grant N00014-01-1-0795, and by AFOSR under grant F49620-02-1-0101.
A preliminary version of this paper appears in {\em Principles of
Knowledge Representation and Reasoning: Proceedings of the Ninth
International Conference (KR 2004)}.}
\\
   Cornell University\\
   Computer Science Department\\
   Ithaca, NY 14853\\
   halpern@cs.cornell.edu\\
   http://www.cs.cornell.edu/home/halpern}

\maketitle

\begin{abstract}
A careful analysis of conditioning in the {\em Sleeping Beauty\/}
problem is done, using the formal model for reasoning about
knowledge and probability developed by Halpern and Tuttle.
While the Sleeping Beauty problem has been viewed as revealing problems
with conditioning in the presence of imperfect recall, the analysis done
here reveals that the problems are not so much due to imperfect
recall as to {\em asynchrony}.  The implications of this 
analysis for van Fraassen's {\em Reflection
Principle\/} and Savage's {\em Sure-Thing Principle} are considered.
\end{abstract}


\section{Introduction}\label{sec:introduction}
The standard approach to updating beliefs in the probability literature
is by conditioning.  But it turns out that conditioning is somewhat
problematic if agents have {\em imperfect recall}.  In the economics
community this issue was brought to the fore by the work of Piccione and
Rubinstein \citeyear{PR97}
(to which was dedicated a special issue of the journal {\em Games and
Economic Behavior}).  There has also been a recent surge of interest in
the topic in the philosophy community, inspired by a re-examination 
by Elga \citeyear{Elga2000} of one of the problems considered by Piccione
and Rubinstein, the so-called {\em Sleeping Beauty problem}.%
\footnote{So named by Robert Stalnaker.}  
(Some recent work on the problem includes
\cite{Art03,Dorr02,Lewis01,Monton02}.) 

I take the Sleeping Beauty problem as my point of departure in this
paper too.  I argue that the problems in updating arise not just with
imperfect recall, but also in {\em asynchronous\/} systems, where agents
do not know exactly what time it is, or do not share a global clock.
Since both human and computer agents are resource bounded and forgetful,
imperfect recall is the norm, rather than an unusual special case.  
Moreover, there are many applications where it is unreasonable to assume
the existence of a global clock.  
Thus, it is important to understand how to do updating in the presence
of asynchrony and imperfect recall.

The Sleeping Beauty
problem is described by Elga as follows:
\begin{quote}
Some researchers are going to put you to sleep.  During the two days
that your sleep will last, they will briefly wake you up either once or
twice, depending on the toss of a fair coin (heads: once; tails: twice).
After each waking, they will put you back to sleep with a drug that
makes you forget that waking.  When you are first awakened, to what
degree ought you believe that the outcome of the coin toss is heads?
\end{quote}
Elga argues that there are two plausible answers.  The first is that it
is $1/2$.  After all, it was $1/2$ before you were put to sleep and you
knew all along that you would be woken up (so you gain no useful
information just by being woken up).  Thus, it should still be
$1/2$ when you are actually woken up.  The second is that it is $1/3$.
Clearly if this experiment is carried out repeatedly, then in the long
run, at roughly one third of the times that you are woken up, you
are in a trial in which the coin lands heads.
   
Elga goes on to give another argument for $1/3$, which he
argues is in fact the correct answer.  Suppose you are put to sleep on
Sunday, so that you are first woken on Monday and then possibly again on
Tuesday if the coin lands tails.  Thus, when you are woken up, there are
three events that you consider possible:
\begin{itemize}
\item $e_1$: it is now Monday and the coin landed heads;
\item $e_2$: it is now Monday and the coin landed tails;
\item $e_3$: it is now Tuesday and the coin landed tails.
\end{itemize}
Elga's argument has two steps:
\begin{enumerate}
\item If, after waking up, you learn that it is Monday, you should
consider $e_1$ and $e_2$ equally likely.  Since, conditional on learning
that it is Monday, you consider $e_1$ and $e_2$ equally likely, you
should consider them equally likely unconditionally.  
\item Conditional 
on the coin landing tails, it also seems reasonable that $e_2$ and $e_3$
should be equally likely; after all, you have no reason to think Monday
is any more or less likely that Tuesday if the coin landed tails.
Thus, unconditionally, $e_2$ and $e_3$ should be equally likely.
\end{enumerate}
From these two steps, it follows that $e_1$, $e_2$, and $e_3$ are
equally likely.  The only way that this could happen is for 
them all to have probability $1/3$.  So heads should have
probability $1/3$.

Suppose that the
story is changed so that (1) heads has probability .99 and tails has
probability .01, (2) you are woken up once if the coin lands heads, and
(3) you are woken up 9900 times if the coin lands tails.  In this case,
Elga's argument would say that the probability of tails is .99.  Thus,
although you know 
you will be woken up whether the coin lands heads or tails, and you are
initially almost certain that the coin will land heads, when you are woken
up (according to Elga's analysis) you are almost certain that the coin
landed tails!  

How reasonable is this argument?  The second step involves an implicit
appeal to the Principle of Indifference.  But note that once 
$e_1$ and $e_2$ are taken to be equally likely, the only way to get
the probability of heads to be $1/2$ is to give $e_3$ probability 0,
which seems quite unreasonable.  Thus, an appeal to the Principle of
Indifference is not critical here to argue that $1/2$ is not the
appropriate answer.  

What about the first step?  If your probability is represented by
$\Pr$ then, by Bayes' Rule,
$$\Pr(\heads \mid \Monday) = \frac{\Pr(\Monday \mid \heads)
\Pr(\heads)}{
\Pr(\Monday \mid \heads)\Pr(\heads) + \Pr(\Monday \mid
\tails)\Pr(\tails)}.
$$
Clearly $\Pr(\Monday \mid \heads) = 1$.  By the Principle of
Indifference, $\Pr(\Monday \mid \tails) = 1/2$.  If we take 
$\Pr(\heads) = \Pr(\tails) = 1/2$, then we get $\Pr(\heads \mid \Monday)
= 2/3$.  Intuitively, it being Monday provides stronger evidence for
heads than for tails, since $\Pr(\Monday \mid \heads)$ is larger than
\mbox{$\Pr(\Monday \mid \tails)$}.  
Of course, this argument already assumes that $\Pr(\heads) = 1/2$, so we
can't use it to argue that $\Pr(\heads) = 1/2$.  The point here is simply
that it is not blatantly obvious that $\Pr(\heads \mid \Monday)$ should
be taken to be $1/2$.%
\footnote{Thanks to Alan H\'ajek for making this point.}

To analyze these arguments, I use a formal model for reasoning
about knowledge and probability that Mark Tuttle and I developed
\cite{HT} (HT from now on), which in turn is based on the ``multiagent
systems'' 
framework for reasoning about knowledge in computing systems, introduced
in \cite{HFfull} (see \cite{FHMV} for motivation and 
discussion).  Using this model, 
I argue that Elga's argument is not as compelling as it may seem,
although not for the reasons discussed above.  The problem turns out to
depend on the difference between the probability of heads conditional on
it being Monday vs.~the probability of heads conditional on {\em
learning\/} that it is Monday.  
The 
analysis also reveals that, despite the focus of the economics community
on imperfect recall, the real problem is not imperfect
recall, but asynchrony:~the fact that Sleeping Beauty does not know
exactly what time it is.

I then consider other arguments and desiderata traditionally
used to justify probabilistic conditioning, such as frequency
arguments, 
betting arguments, van Fraassen's \citeyear{vanF84} {\em Reflection
Principle}, and Savage's \citeyear{Savage} {\em Sure-Thing
Principle}. I show that our intuitions for these arguments are  
intimately bound up with assumptions such as synchrony and perfect
recall.

The rest of this paper is organized as follows.  In the next section 
I review the basic multiagent systems framework.  In Section 3, I describe
the HT approach to adding probability to the framework when the system
is synchronous.  
HT generalized their approach to the asynchronous case; their generalization
supports the ``evidential argument'' above, giving the answer
$1/2$  in the Sleeping Beauty problem.
I also consider a second generalization, which gives the answer $1/3$ 
in the Sleeping Beauty problem (although not exactly by Elga's reasoning).
In Section~\ref{sec:compare}, I consider other arguments and desiderata.
I conclude in Section~\ref{sec:conc}.

\section{The framework}\label{sec:framework}

\subsection{The basic multiagent systems framework}
In this section, we briefly review the multiagent systems framework; 
see \cite{FHMV} for more details.

A {\em multiagent system\/} consists of $n$ agents interacting over
time.  At each point in time, each agent is in some 
{\em local state\/}. Intuitively, an agent's
local state encapsulates all the information to which the agent has
access.  For example, 
in a poker game, a player's state might consist of the cards he
currently holds, the bets made by the other players,
any other cards he has seen, and any information he may have about the
strategies of the other players (e.g., Bob may know that Alice
likes to bluff, while Charlie tends to bet conservatively).
In the Sleeping Beauty problem, we can assume
that the agent has local states corresponding to 
``just woken up'', ``just before the experiment'', and ``just
after the experiment''. 

Besides the agents, it is also conceptually useful to have an
``environment'' (or ``nature'') whose state can be thought of as
encoding everything relevant to the description of the system that may
not be included in the agents' local states.  
For example, in the Sleeping Beauty problem, the
environment state can encode the actual day of the week and the outcome
of the coin toss.  In many ways, the environment can be viewed as just
another agent. In fact, in the case of the Sleeping Beauty problem, the
environment can be viewed as the local state of the experimenter.

We can view the whole system as being in some {\em
global state}, a tuple consisting of the local state of each agent and the
state of the environment. 
Thus, a global state has the form $(s_e, s_1,\ldots, s_n)$,
where $s_e$ is the state of the environment and $s_i$ is agent $i$'s
state,
for $i = 1, \ldots , n$. 

A global state describes the system at a given point in time.  But
a system is not a static entity.  It is constantly changing over time.
A {\em run\/}\index{run|(} captures the dynamic aspects of a system.
Intuitively, a run is a complete description of one possible way in
which the system's state can evolve over time.  Formally, a
run is a function from time to global states.
For definiteness, I take time to range over the natural numbers.
Thus, $r(0)$ describes the initial global state of the system in a
possible execution, $r(1)$ describes the next global state, and so on.
A pair $(r,m)$ consisting of a run $r$ and time $m$ is called a {\em
point}.\index{point}
If $r(m) = (s_e, s_1, \ldots, s_n),$ then define $r_e(m) = s_e$ and
$r_i(m) = s_i,$ $i = 1, \ldots, n$; thus, $r_i(m)$ is agent $i$'s
local state at the point $(r,m)$ and $r_e(m)$ is the environment's state
at $(r,m)$.  I write $(r,m) \sim_i (r',m')$ if agent $i$ has the same
local state at both $(r,m)$ and $(r',m')$, that is, if $r_i(m) = r'_i(m')$.
Let $\K_i(r,m) = \{(r',m') : (r,m) \sim_i  (r',m')\}$.  
Intuitively, $\K_i(r,m)$ is the set of points that $i$ considers
possible at $(r,m)$;
these are the states that $i$ cannot distinguish based basis of $i$'s
information at $(r,m)$.
Sets of the form $\K_i(r,m)$ are sometimes called {\em
information sets}.

In general, there are many possible executions of a system:\ there could
be a number of possible initial states and many things that could happen from
each initial state. For example, in a draw poker game, the initial
global states 
could describe the possible deals of the hand by having player $i$'s
local state describe the cards held by player $i$.  For each fixed
deal of the cards, there may still be many
possible betting sequences, and thus many runs.
Formally, a {\em system\/}\index{multiagent system} is a nonempty set of runs.
Intuitively, these runs describe all the possible sequences of events
that could occur in the system.  Thus, I am essentially identifying a
system with its possible behaviors.\index{state!global|)}  

There are a number of ways of modeling the Sleeping Beauty problem as a
system.  Perhaps simplest is to consider it as a single-agent problem,
since the experimenter plays no real role.  (Note that it is important
to have the environment though.)
Assume for now that the system modeling the Sleeping Beauty problem
consists of two runs, 
the first corresponding to the coin landing heads, and the second
corresponding to the coin landing tails.   (As we shall see, while
restricting to two runs seems reasonable, it may not capture all aspects
of the problem.)
There are still some choices to be
made with regard to modeling the global states.  Here is one way:
At time 0, a coin is tossed; the environment state encodes the outcome.
At time 1, the agent
is asleep (and thus is in a ``sleeping'' state).  At time 2, the agent
is woken up.  If the coin lands tails, then at time 3, the agent is back
asleep, and at time 4, is woken up again.  
Note that I have
assumed here that time in both of these runs ranges from 0 to 5.
Nothing would change if I allowed runs to have infinite length or
a different (but sufficiently long) finite length.

\begin{figure}[htb]
\centerline{\includegraphics[height=1.8in]{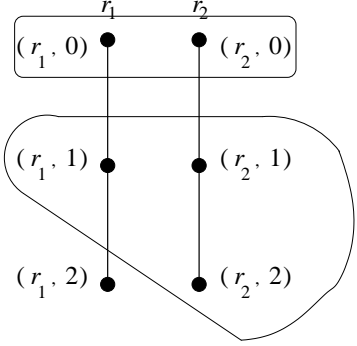}}
\caption{The Sleeping Beauty problem, captured using
$\R_1$.}\label{fig:sb1}
\end{figure}

Alternatively, we might decide that it is not important to model the
time that the agent is sleeping; all that matters is the point
just before the agent is put to sleep and the points where the agent is
awake.  Assume that Sleeping Beauty is in
state $b$ before the experiment starts, in state $a$ after
the experiment is over, and in state $w$ when
woken up.  This leads to a model with two runs $r_1$ and $r_2$, where 
the first three global states in $r_1$ are $(H,b)$, $(H,w)$, and
$(H,a)$, and the first four global states in $r_2$ are $(T,b)$, $(T,w)$,
$(T,w)$, $(T,a)$.  
Let $\R_1$ be the system consisting of the runs $r_1$ and $r_2$.  This
system is  shown in Figure~\ref{fig:sb1} 
(where only the first three global states in each run are shown).
The three points where the
agent's local state is $w$, namely, $(r_1, 1)$, $(r_2, 1)$, and $(r_2,
2)$, form 
what is traditionally called in game theory an {\em information set}.
These are the three points that the agent considers possible when she is
woken up.  For definiteness, I use $\R_1$ in much of my analysis of
Sleeping Beauty.  

Notice that $\R_1$ is also compatible with a somewhat
different story.  Suppose that the agent is not aware of time passing.  At
time 0 the coin is tossed, and the agent knows this.  If the coin lands
heads, only one round passes before the agent is told that the
experiment is over; if the 
coin lands tails, she is told after two rounds.  Since the agent is not
aware of 
time passing, her local state is the same at the points $(r_1,2)$,
$(r_2,1)$, and $(r_2,2)$.  The same analysis should apply to the
question of what the probability of heads is at the information set. 
The key point is that here the agent does not forget; she is simply
unaware of the time passing.

Various other models are possible:
\begin{itemize}
\item We could assume (as Elga does at one point) that the coin
toss happens only after the agent is woken up the first time.  Very
little would change, except that the environment state would be
$\emptyset$ (or some other way of denoting that the coin hasn't been
tossed) in the first two global states of both runs. 
Call the two resulting runs $r_1'$ and $r_2'$.

\item All this assumes that the agent knows when the coin is going to be
tossed.  If the agent doesn't know this, then we can consider the system
consisting of the four runs $r_1, r_1', r_2, r_2'$.  

\item 
Suppose that we now want to allow for the possibility that, upon
wakening, the agent learns that it is Monday (as in Elga's argument).
To do this, the system must include runs where the agent actually learns
that it is Monday.  
Now two runs no longer suffice.
For example,
we can consider the system $\R_2 = (r_1, r_2, r_1^*, r_2^*)$, where
$r_i^*$ is the 
same as $r_i$ except that on Monday, the agent's local state encodes that
it is Monday.  Thus, the sequence of global states in $r_1^*$ is
$(H,b)$, $(H,M)$, $(H,a)$, and the sequence in $r_2^*$ is
$(T,b)$, $(T,M)$, $(T,w)$.  $\R_2$ is
described in Figure~\ref{fig:sb2}.
Note that on Tuesday in $r_2^*$, the agent forgets whether she was woken
up on Monday.  She is in the same local state on Tuesday in $r_2^*$ as
she is on both Monday and Tuesday in $r_2$.  
\begin{figure}[htb]
\centerline{\includegraphics[height=1.8in]{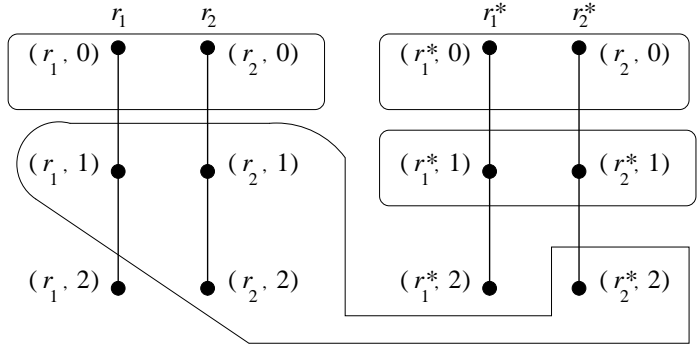}}
\caption{An alternate representation of the Sleeping Beauty problem,
using $\R_2$.}
\label{fig:sb2}
\end{figure}
\end{itemize}

Yet other representations of the Sleeping Beauty problem are
also possible.  The point that I want to emphasize here is that the
framework has the resources to capture important distinctions about when
the coin is tossed and what agents know.

\subsection{Synchrony and perfect recall}
One advantage of the multiagent systems framework is that it can be used
to easily model a number of important assumptions.  I focus on two of
them here: {\em synchrony}, the assumption that agents know the
time, and {\em perfect recall}, the assumption that agents do not
forget \cite{FHMV,HV2}

Formally, a system $\R$ is
{\em synchronous for agent $i$\/}\index{synchronous system}
if for all points $(r,m)$
and $(r',m')$ in~$\R,$ if $(r,m) \sim_i (r',m'),$ then $m=m'$.  Thus, if
$\R$ is synchronous for agent $i,$ then at 
time $m,$ agent~$i$ knows that it is time $m,$
because it is time $m$ at all the points he considers possible.
$\R$ is {\em synchronous\/} if it is synchronous for all agents.
Note that the systems that model the Sleeping Beauty problem are not
synchronous.  When Sleeping Beauty is woken up on Monday, she does not know
what day it is.

Consider the following example of a synchronous system, taken from
\cite{Hal31}: 
\xam\label{pchdyn.xam1}
Suppose that Alice tosses two coins and sees how the coins land.  Bob
learns how the first coin landed after the second coin is tossed, but
does not learn the outcome of the second coin toss.  How should this be
represented as a 
multiagent system?  The first step is to decide what the local states
look like.  There is no ``right'' way of modeling the local states.
What I am about to describe is one reasonable way of doing it, but
clearly there are others.

The environment state will be used to model what actually happens.
At time 0, it is $\< \, \>,$ the empty sequence, indicating that nothing
has yet happened.  At time 1, it is either $\<H\>$ or $\<T\>,$ depending
on the outcome of the first coin toss.  At time 2, it is either
$\<H,H\>,$ $\<H,T\>,$ 
$\<T,H\>,$ or $\<T,T\>,$ depending on the outcome of both coin tosses.
Note that the environment state is
characterized by the values of two random variables, describing the
outcome of each 
coin toss.  Since Alice knows the outcome of the coin tosses, 
I take Alice's local state to be the same as the environment state at all
times.  

What about Bob's local state?  After the first coin is tossed, Bob still
knows nothing; he learns the outcome of the first coin toss after the
second coin is tossed.  The first thought might then be to take his
local states to have the form $\< \, \>$ at time 0 and time 1 (since he 
does not know the outcome of the first coin toss at time 1) and either
$\<H\>$ or $\<T\>$ at time 2.  This choice would not make the system
synchronous, since Bob would not be able to distinguish time 0 from time 1.
If Bob is aware of the passage of time, then at
time 1, Bob's state must somehow encode the fact that the time is 1.  I
do this by taking Bob's state at time 1 to be $\<\tick\>,$ to denote
that one time tick has passed.  (Other ways of encoding the time are, of
course, also possible.)  Note that the time is already implicitly
encoded in Alice's state: the time is 1 if and only if her state is
either $\<H\>$ or $\<T\>$.

Under this representation of global states, there are seven
possible global states:
\begin{itemize}
\item $(\<\, \>, \<\, \>, \<\,\>),$ the initial state,
\item two time-1 states of the form
$(\<X_1\>, \< X_1 \>, \<\tick\>),$ for $X_1 \in \{H,T\}$,
\item four time-2 states of the form
$(\<X_1,X_2\>, \< X_1,X_2\>, \<\tick,X_1\>),$ for $X_1, X_2 \in
\{H,T\}$.
\end{itemize}
In this simple case, the environment state determines the global state
(and is identical to Alice's state),
but this is not always so.

The system describing this situation has four runs, $r^1, \ldots, r^4,$
one for each of the
time-2 global states.  The runs are perhaps best thought of as being the
branches
of the computation tree described in
Figure~\ref{pchdyn.fig1}.
\medskip

\setlength{\unitlength}{.3in}
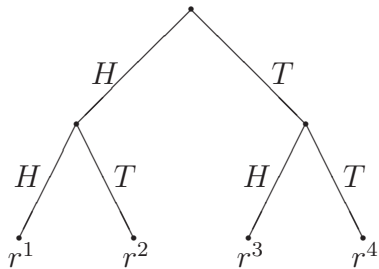
\begin{figure}[htb]
\begin{center}
\begin{picture}(8,4)(0,1.5)
\multiput(1,2)(2,0){4}{\circle*{.1}}
\multiput(2,4)(4,0){2}{\circle*{.1}}
\multiput(4,6)(8,0){1}{\circle*{.1}}
\multiput(2,4)(4,0){2}{\line(-1,-2){1}}
\multiput(2,4)(4,0){2}{\line(1,-2){1}}
\multiput(4,6)(8,0){1}{\line(-1,-1){2}}
\multiput(4,6)(8,0){1}{\line(1,-1){2}}
\put(2.25,4.7){$H$}
\put(5.4,4.7){$T$}
\put(0.9,2.9){$H$}
\put(2.65,2.9){$T$}
\put(4.9,2.9){$H$}
\put(6.65,2.9){$T$}
\put(0.8,1.5){$r^1$}
\put(2.8,1.5){$r^2$}
\put(4.8,1.5){$r^3$}
\put(6.8,1.5){$r^4$}
\end{picture}
\caption{Tossing two coins.}
\label{pchdyn.fig1}
\end{center}
\end{figure}

\exam

Modeling perfect recall in the systems framework 
requires a little care.  In this framework, an agent's
knowledge is determined by his local state.  Intuitively, an agent has
perfect recall if his local state is always ``growing'', by adding the
new information he acquires over time.  
This is essentially how
the local states were modeled in Example~\ref{pchdyn.xam1}.
In general, local states are not required to grow in this sense, quite
intentionally.  It is quite possible that information
encoded in~$r_i(m)$---$i$'s local state at time~$m$ in run~$r$---no
longer appears in~$r_i(m+1)$.   Intuitively, this means that agent~$i$
has lost or ``forgotten'' this information.  
There are often scenarios of interest where
it is important to model the fact that certain information is discarded.
In practice, for example, an agent may simply not have enough
memory capacity to remember everything he has learned.
Nevertheless, although
perfect recall is a strong assumption,
there are many instances where it is natural to model agents
as if they do not forget.

Intuitively, an agent with perfect recall should be
able to reconstruct his complete local history from his current
local state.  
To capture this intuition,
let {\em agent~$i$'s local-state sequence at the point
$(r,m)$\/}\index{local-state sequence}
be the sequence of local states that she has gone through in run~$r$ up
to time~$m,$ without consecutive repetitions.  Thus, if from time 0
through time~4 in run~$r$ agent~$i$
has gone through the sequence $\< s_i,s_i,s_i',s_i,s_i \>$ of
local states, where $s_i \ne s_i',$
then her local-state sequence at $(r,4)$ is $\< s_i,s_i',s_i \>$.
Agent~$i$'s local-state sequence at a point $(r,m)$ essentially
describes what has happened in the run up to time~$m,$ from~$i$'s point
of view. Omitting consecutive repetitions is intended to capture
situations where the agent has perfect recall but 
is not aware of time passing, so she cannot
distinguish a run where she stays in a given state $s$ for three rounds
from one where she stays in $s$ for only one round.

\begin{figure}[htb]
\centerline{\includegraphics[width=2in]{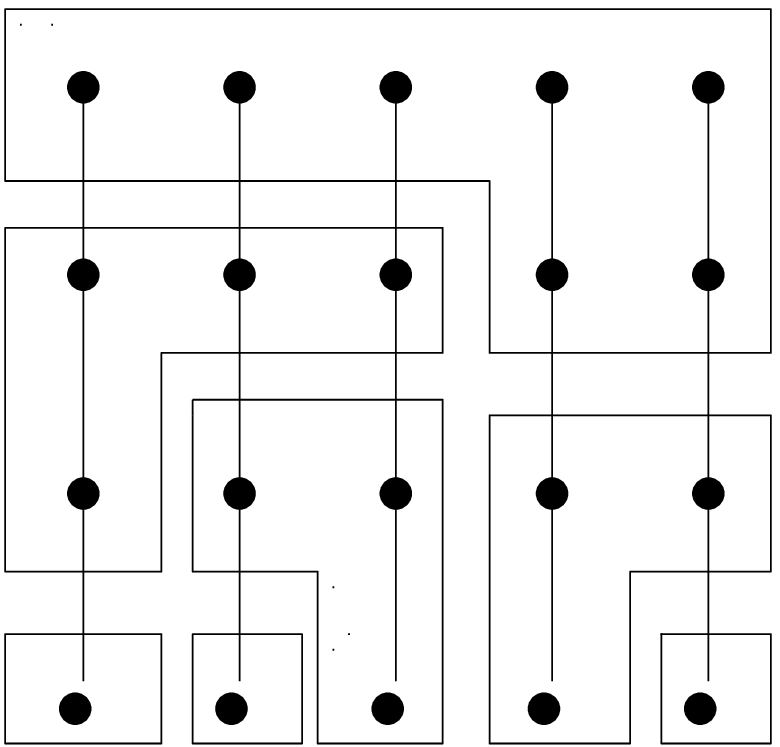}}
\caption{An asynchronous system where agent $ i $ has perfect recall.}
\label{pchdyn.fig3}
\end{figure}

An agent has perfect recall if her current local state
encodes her whole local-state sequence.  More formally,
{\em agent~$i$ has perfect recall in system~$\R$\/}\index{perfect recall}
if, at all points $(r,m)$ and $(r',m')$ in~$\R,$
if $(r,m) \sim_i (r',m'),$ then
agent~$i$ has the same local-state sequence at both $(r,m)$ and
$(r',m')$.  Thus,
agent~$i$ has perfect recall if she ``remembers'' her local-state
sequence at all times.%
\footnote{This definition of perfect recall is not quite the same as that used
in the game theory literature, where agents must
explicitly recall the actions taken  (see \cite{Hal15} for a discussion
of the issues),
but the difference between the two notions is not relevant here.
In particular, according to both definitions,
the agent has perfect
recall in the ``game'' described by Figure~\ref{fig:sb1}.}
In a system with perfect recall, $r_i(m)$ encodes~$i$'s
local-state sequence in that, at
all points where~$i$'s local state is $r_i(m),$
she has the same local-state sequence.
A system where agent $ i $ has perfect recall is shown in
Figure~\ref{pchdyn.fig3}.

\begin{figure}[htb]
\centerline{\includegraphics[height=2.1in]{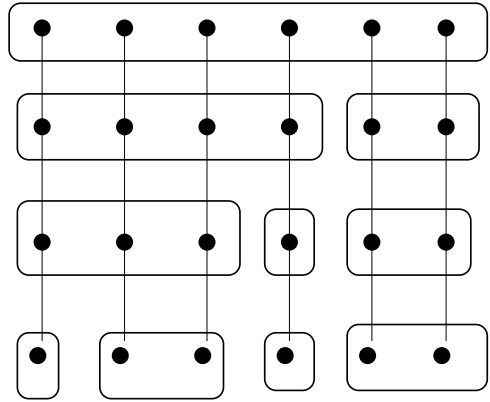}}
\caption{A synchronous system with perfect recall.}
\label{fig:sync}
\end{figure}

The combination of synchrony and perfect recall leads to particularly 
pleasant properties.  It is easy to see
that if $\R$ is a synchronous system with perfect recall and $(r,m+1)
\sim_i (r',m+1)$, then $(r,m) \sim_i (r',m)$.  That is, if agent $i$
considers run $r'$ possible at the point $(r,m+1)$, then 
$i$ must
also consider run $r'$ possible at the point $(r,m)$.  (Proof: since the
system is synchronous and $i$ has perfect recall, 
$i$'s local state must be different at each point in $r$.  For if $i$'s
local state were the same at two points $(r,k)$ and $(r,k')$ for $k \ne
k'$, then agent $i$ would not know that it was time $k$ at the point
$(r,k)$.  Thus, at the points $(r,m+1)$,  $i$'s local-state
sequence must have length $m+1$.  Since $(r,m+1) \sim_i (r',m+1)$, $i$
has the same local-state sequence at $(r,m+1)$ and $(r',m+1)$.  Thus, $i$
must also have the same local-state sequence at the points $(r,m)$ and
$(r',m)$, since $i$'s local-state sequence at these points is just the
prefix of $i$'s local-state sequence at $(r,m+1)$ of length $m$.  
It is then immediate that $(r,m) \sim_i (r',m)$.)
Thus, in a synchronous system with perfect recall, agent $i$'s
information set refines over time,
as shown in Figure~\ref{fig:sync}.%
\footnote{In the language of probabilists, in synchronous systems with
perfect recall, information sets form a {\em filtration} \cite[Section
35]{Billingsley}.  The importance of assuming that the information sets
form a filtration in the context of the Sleeping Beauty problem is
emphasized by Schervish, Seidenfeld, and Kadane \citeyear{SSK04}.
However, my analysis applies in the asynchronous case applies despite the
fact that the information sets do not form a filtration.}

Note that whether the agent has perfect recall in the Sleeping Beauty
problem depends in part on how we model the problem.  In the system
$\R_1$ she does; in 
$\R_2$ she does not.  For example, at the point
$(r_2^*,2)$ in $\R_2$, where her local state is $(T,w)$, she has
forgotten that she was woken up at time 1 (because she cannot
distinguish $(r_2,2)$ from $(r_2^*,2)$).  (It may seem strange that the
agent has perfect recall in $\R_1$, but that is because in $\R_1$, the
time that the agent is asleep is not actually modeled.  It happens
``between the points''.  If we explicitly include local states where the
agent is asleep, then the agent would not have perfect recall in the
resulting model.   The second interpretation of $\R_1$, where the
agent is unaware of time passing, is perhaps more compatible with perfect
recall.  I use $\R_1$ here so as to stress that perfect recall is not
really the issue in the Sleeping Beauty problem; it is the asynchrony.)

\section{Adding probability}

To add probability to the framework, I start by assuming a probability
on the set of runs in a system.  Intuitively, this should be thought of
as 
the agents' common probability.  
It is not necessary to assume that the
agents all have the same probability on runs; different agents may have
use probability measures.  Moreover, it is not necessary to assume that
the probability is placed on the whole set of runs.  There are many
cases where it is convenient to partition the set of runs and put a
separate probability measure on each cell in the partition (see
\cite{Hal31} for a discussion of these issues).  However, to analyze the
Sleeping Beauty problem, it suffices to have a single probability on
the runs.  
A {\em probabilistic system\/} is a pair
$(\R,\Pr)$, where $\R$ is a system (a set of runs) and $\Pr$ is a
probability on $\R$.  (For simplicity, I assume that $\R$ is finite and
that all subsets of $\R$ are measurable.)  In the case of the Sleeping Beauty
problem, the probability on $\R_1$ is immediate from the description of
the problem: each of $r_1$ and $r_2$ should get probability $1/2$.
To determine a probability on the runs of $\R_2$, we need to
decide how likely it is that the agent will discover that it is actually
Monday.  Suppose that probability is $\alpha$.  In that case, $r_1$
and $r_2$ both get probability $(1- \alpha)/2$, while $r_1^*$
and $r_2^*$ both get probability $\alpha/2$.  

Unfortunately, the probability on runs is not enough for the agent to
answer questions like ``What is the probability that heads was
tossed?''~if she is asked this question at the point $(r_1,1)$ when she
is woken up in $\R_1$, for example.  At this point she considers three
points possible: $(r_1,1)$, $(r_2, 1)$, and $(r_2, 2)$, the three points
where she is woken up.  She needs to put a probability on this space of
three points to answer the question.   Obviously, the probability on the
points should be related to the probability on runs.  But how?  That is
the topic of this section.

As the preceding discussion should make clear, 
points can be viewed possible worlds.  In HT,
a modal logics of knowledge and probability is considered where truth is
defined relative to points in a system.  Points are 
somewhat analogous to what Lewis \citeyear{Lewis79} calls {\em
centered\/} possible worlds, since they are equipped with a time
(although they are not equipped with a designated individual).  Runs can
then be viewed as uncentered possible worlds.   Lewis~\citeyear{Lewis79}
argued that credence should be placed not on possible worlds, but on
centered possible worlds.  The key issue here is that in many
applications, it is more natural to start with a probability on
uncentered worlds; the question is how to define a probability on
centered worlds.%
\footnote{As a cultural matter, in the computer science literature,
defining truth/credence relative to centered worlds is the norm.
Computer scientists are, for example, interested in temporal logic for
reasoning about what happens while a program is running \cite{MP1}.
Making time part of the world is necessary for this reasoning.
Interestingly, economists, like philosophers, have tended to focus on
uncentered worlds.
I have argued elsewhere \cite{Hal15} 
that centered worlds (represented as points) are necessary to capture
some important temporal considerations in the analysis of games.}

\subsection{The synchronous case}
Tuttle and I suggested a relatively straightforward
way of going from a probability on runs to a probability on points in
synchronous systems. For all times $m$, the probability $\Pr$ on $\R$, the set
of runs, can be used to put a probability $\Pr^m$ on the points in  $\R^m
= \{(r,m): r \in \R\}$:~simply take $\Pr^m(r,m) = \Pr(r)$.  
Thus, the probability of the point $(r,m)$ is just the probability of
the run $r$.  Clearly, $\Pr^m$ is a well-defined probability on the set
of time-$m$ points.
Since $\R$
is synchronous, at the point $(r,m)$, agent $i$ considers possible 
only time-$m$ points.  That is, all the points 
in $\K_i(r,m) = \{(r',m'):  (r,m) \sim_i (r',m')\}$ are actually
time-$m$ points.  
Since, at the point $(r,m)$, the agent considers possible only the
points in $\K_i(r,m)$, it seems reasonable to take the agent's
probability at the point $(r,m)$ to the result of conditioning $\Pr^m$
on $\K_i(r,m)$, provided that $\Pr^m(\K_i(r,m)) > 0$, which, for
simplicity, I assume here.  Taking ${\Pr}_{(r,m,i)}$ to denote agent $i$'s 
probability at the point $(r,m)$, this suggests that ${\Pr}_{(r,m,i)}(r',m)
= \Pr^m((r',m) \mid \K_i(r,m))$.

To see how this works, consider the system of 
Example~\ref{pchdyn.xam1}.  Suppose that 
the first coin has
bias $2/3$, the second coin is fair, and the coin tosses are
independent, as shown in Figure~\ref{pchdyn.fig2}.
Note that, in Figure~\ref{pchdyn.fig2}, the edges coming out of each
node are labeled with a probability, which is intuitively the
probability of taking that transition.  Of course, the probabilities
labeling the edges coming out of any fixed node must sum to 1, since
some transition must be taken. For example, the edges coming out of the
root have probability $2/3$ and $1/3$.   Since the transitions in this
case (i.e., the coin tosses) are assumed to be independent, it is
easy to compute the probability of each run.  For example, the
probability of run $r^1$ is $2/3 \times 1/2 = 1/3$; this represents the
probability of getting two heads.

\setlength{\unitlength}{.3in}
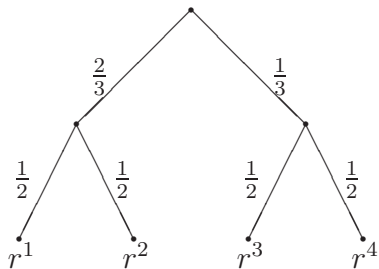
\begin{figure}[htb]
\begin{center}
\begin{picture}(8,4)(0,1)
\multiput(1,1)(2,0){4}{\circle*{.1}}
\multiput(2,3)(4,0){2}{\circle*{.1}}
\multiput(4,5)(8,0){1}{\circle*{.1}}
\multiput(2,3)(4,0){2}{\line(-1,-2){1}}
\multiput(2,3)(4,0){2}{\line(1,-2){1}}
\multiput(4,5)(8,0){1}{\line(-1,-1){2}}
\multiput(4,5)(8,0){1}{\line(1,-1){2}}
\put(2.25,3.7){$\frac{2}{3}$}
\put(5.4,3.7){$\frac{1}{3}$}
\put(0.9,1.9){$\frac{1}{2}$}
\put(2.65,1.9){$\frac{1}{2}$}
\put(4.9,1.9){$\frac{1}{2}$}
\put(6.65,1.9){$\frac{1}{2}$}
\put(0.8,0.5){$r^1$}
\put(2.8,0.5){$r^2$}
\put(4.8,0.5){$r^3$}
\put(6.8,0.5){$r^4$}
\end{picture}
\caption{Tossing two coins, with probabilities.}
\label{pchdyn.fig2}
\end{center}
\end{figure}

\subsection{The general case}\label{sec:general}

The question now is how the agents should ascribe probabilities in 
arbitrary (not necessarily synchronous) system, such as that of the
Sleeping Beauty problem. 
The approach suggested above does not immediately extend to the
asynchronous case.  In the asynchronous case, the points in $\K_i(r,m)$
are not in general all time-$m$ points, so it does not make sense to
condition $\Pr^m$ on $\K_i(r,m)$.  (Of course, it would be possible
to condition on the time-$m$ points in $\K_i(r,m)$, but it is easy to
give examples showing that doing this gives rather nonintuitive results.)

I discuss two reasonable candidates for ascribing probability in the
asynchronous case here, which are 
generalizations of the two approaches that Elga considers. 
I first consider these approaches in the context of the Sleeping
Beauty problem, and then give the general formalization.  

Consider the
system described in 
Figure~\ref{fig:sb1}, but now suppose that the probability of $r_1$ is
$\beta$ and the probability of $r_2$ is $1-\beta$.  (In the original
Sleeping Beauty problem, $\beta = 1/2$.)  It seems reasonable that at
the points $(r_1,0)$ and $(r_2,0)$, the agent ascribes probability 
$\beta$ to $(r_1,0)$ and $1-\beta$ to $(r_2,0)$, using the HT approach
for the synchronous case.  What about at each of the points $(r_1,1)$,
$(r_2, 1)$, and $(r_2,2)$?  One approach (which I henceforth call the
{\em HT approach}, since it was advocated in HT), is to say
that the probability $\beta$ of run $r_1$ is projected to the point
$(r_1,1)$, while the probability $1-\beta$ of $r_2$ is projected to
$(r_2,1)$ and $(r_2,2)$. How should the probability be split over these
two points?  Note that splitting the probability essentially 
amounts to deciding the relative probability of being at time 1 and time
2.  Nothing in the problem description gives us any indication of how to
determine this.  HT avoid making this determination by making
the singleton sets $\{(r_2,1\}\}$ and $\{(r_2,2)\}$ nonmeasurable.  
Since they are not in the domain of the probability measure, there is no
need to give them a probability.
The only measurable sets in this 
space would then be $\emptyset$, 
$\{(r_1,1)\}$, $\{(r_2,1),(r_2,2)\}$, and $\{(r_1,1), (r_2,1),
(r_2,2)\}$, which get probability 0, $\beta$, $1-\beta$, and 1,
respectively. 
An alternative is to 
apply the Principle of Indifference and
take times 1 and 2 to be equally likely.  In this
case the probability of the set $\{((r_2,1),(r_2,2)\}$ is split over
$(r_2,1)$ and $(r_2,2)$, and they each get probability $(1-\beta)/2$.
When $\beta = 1/2$, this gives Elga's first solution.
Although it is reasonable to assume that times 1 and 2 are equally
likely, the technical results that I prove hold no matter how the
probability is split between times 1 and 2.

The second approach, which I call the {\em Elga approach} (since it turns out
to generalize what Elga does), is to require that for any pair of points
$(r,m)$ and $(r',m')$ on different runs, the relative probability of
these points is the same as the relative probability of $r$ and $r'$.  
This property is easily seen to hold for the HT approach in the
synchronous case.  With this approach, the ratio of the probability of
$(r_1,1)$ and $(r_2,1)$ is $\beta: 1-\beta$, as is the ratio of the
probability of $(r_1,1)$ and $(r_2,2)$.  This forces the probability of
$(r_1,1)$ to be $\beta/(2 - \beta)$, and the probability of each
of $(r_1,1)$ and $(r_2,2)$ to be $(1-\beta)/(2-\beta)$.  
Note that, according to the Elga approach, if $\Pr$ is the
probability on the runs of $\R_1$,  $\beta = 1/2$, 
so that $\Pr(r_1) = \Pr(r_2) = 1/2$, 
and $\Pr'$ is the probability that the agent assigns to the three points
in the information set, then 
$$\begin{array}{ll}
&{\Pr}'((r_1,1) \mid \{(r_1,1), (r_2,1)\})\\
= &{\Pr}'((r_1,1) \mid \{(r_1,1), (r_2,2)\})\\ 
= &\Pr(r_1 \mid \{r_1,r_2\})\\
= &1/2.
\end{array}$$ 
Thus, we must have $\Pr'((r_1,1)) = \Pr'((r_2,1)) = \Pr'((r_2,2))$, so
each of the three points has probability $1/3$, which is Elga's second
solution.  Moreover, note that 
$$
{\Pr}'((r_1,1) \mid \{(r_1,1), (r_2,1)\})
= {\Pr}'((r_2,1) \mid \{(r_1,1), (r_2,2)\})
= 1/2.
$$
This is one way of formalizing the first step of Elga's argument; i.e., 
that $\Pr'$ should have the property that, conditional on learning it is
Monday, you should consider ``it is now Monday and
the coin landed heads'' and ``it is now Monday and the coin landed tails''
equally likely.  
The second step of Elga's argument used the Principle of Indifference to
conclude that, if the coin landed tails, then all days were equally
likely.  That use of the Principle of Indifference is implicit in the
assumption that the relative probability of $(r_1,m)$ and $(r_2,m)$ is the
same for $m=1$ and $m=2$.

To summarize, the HT approach assigns probability among points 
in an information set $I$ by dividing the probability of a  run
$r$ among the points in $I$ that lie on $r$ (and then normalizing so
that the sum is one), while the Elga approach proceeds by giving each and
every point in $I$ that is on run $r$  the same probability as that of
$r$, and then normalizing. 

For future reference, I now give a somewhat more precise formalization
of the HT and Elga approaches.  To do so, 
it is helpful to have some notation that relates sets of runs to sets of
points.  If $\S$ is a set of runs and $U$ is a set of points, let
$\S(U)$ be the set of runs in $\S$ going through some point in $U$. 
and let $U(\S)$ be the set of 
points in $U$ that lie on some run in $\S$.  That is,
$$\begin{array}{c}
\S(U) = \{r \in \S:  (r,m) \in U \mbox{ for some $m\}$ and}\\
U(\S) = \{(r,m) \in U : r \in \S\}.
\end{array}$$
Note that, in particular, $\K_i(r,m)(r')$ is the set of points in the
information set $\K_i(r,m)$ that are on the run $r'$ and $\R(\K_i(r,m))$
is the set of runs in the system $\R$ that contain points in
$\K_i(r,m)$.  According to the HT approach, if ${\Pr}_i$ is agent
$i$'s probability on $\R$, the set of runs, then
${\Pr}_{(r,m,i)}^{{\it HT}}(\K_i(r,m)(r')) = {\Pr}_i(r' \mid
\R(\K_i(r,m)))$.
(Note  that here I am using ${\Pr}_{(i,r,m)}^{{\it HT}}$ to denote agent
$i$'s probability at the point $(r,m)$ calculated using the HT approach;
I similarly will use ${\Pr}_{(i,r,m)}^{{\it Elga}}$ to denote agent $i$'s
probability calculated using the Elga approach.)
That is, the probability that agent $i$ assigns at the point $(r,m)$ to
the points in $r'$ is just the probability of the run $r'$ conditional
on the probability of the runs going through the information set
$\K_i(r,m)$.  As I said earlier, Halpern and Tuttle do not try to
assign a probability to individual  points in $\K_i(r,m)(r')$ if there
is more than one point on $r'$ in $\K_i(r,m)$.

By way of contrast, the Elga approach is defined as follows:
$${\Pr}_{(r,m,i)}^{{\it Elga}}(r',m') = \frac{{\Pr}_i(\{r'\} \inter
\R(\K_i(r,m)))}{\sum_{r'' \in \R(\K_i(r,m))} {\Pr}_i(r'')|\K_i(r,m)(\{r''\})|}.$$
It is easy to check that ${\Pr}_{(r,m,i)}^{{\it Elga}}$ is the unique
probability measure $\Pr'$ on $\K_i(r,m)$ such that
$\Pr'((r_1,m_1))/\Pr'((r_2,m_2)) = {\Pr}_i(r_1)/{\Pr}_i(r_2)$ if ${\Pr}_i(r_2)
> 0$.  Note that ${\Pr}_{(r,m,i)}^{{\it Elga}}$ assigns equal probability
to all points on a run $r'$ in $\K_i(r,m)$.  Even if 
${\Pr}_{(r,m,i)}^{{\it HT}}$ is extended so that all points on a given run
are taken to be equally likely, in general, ${\Pr}_{(r,m,i)}^{{\it HT}}
\ne {\Pr}_{(r,m,i)}^{{\it Elga}}$.  The following lemma characterizes
exactly when the approaches give identical results.

\lem\label{lem:identical} ${\Pr}_{(r,m,i)}^{{\it Elga}} =
{\Pr}_{(r,m,i)}^{{\it HT}}$ 
iff $|\K_i(r,m)(\{r_1\})| = |\K_i(r,m)(\{r_2\})|$ for all
runs $r_1, r_2 \in \R(\K_i(r,m))$ such that ${\Pr}_i(r_j) \ne 0 $ for $j =
1, 2$.  
\elem

Note that, in the synchronous case, 
$|\K_i(r,m)(\{r'\})| = 1$ for all runs $r' \in \R(\K_i(r,m))$, so the
two approaches are guaranteed to give the same answers.

\section{Comparing the Approaches}\label{sec:compare}

I have formalized two approaches for ascribing probability in
asynchronous settings, both of which generalize the relatively
noncontroversial approach used in the synchronous case.
Which is the most appropriate?  I examine a number of arguments here.

\subsection{Elga's Argument}
Elga argued for the Elga approach, using the argument that if you
discover or learn that it is Monday, then you should consider heads and tails
equally likely.  As I suggested above, I do not find this a compelling
argument for the Elga approach.  I agree that if you learn that it is
Monday, you should consider heads and tails equally likely.  On the
other hand,  
Sleeping Beauty does not actually learn that it is Monday.  Elga is
identifying the probability of heads conditional on learning that it is
Monday with the probability of heads given that it is Monday.
While these probabilities could be equal, they certainly do not have to
be. An example of Thomason makes the point nicely:%
\footnote{Thanks to Jim Joyce for pointing out this example.}
If I think my wife is much more clever than I, then
I might be convinced that I will never learn of her infidelity should
she be unfaithful.  So, my conditional probability for $Y$, ``I will
learn that my wife is cheating on me'', given $X$,``She will cheat
on me'', is very low.  Yet, the probability of $Y$ if I actually learn $X$
is clearly 1.%
\footnote{There are other reasons why the probability of $Y$ given $X$
might be different from the probability of $Y$ given that you learn or
observe $X$.  In the latter case, you must take into account how you came
to learn that $X$ is the case.  Without taking this into account, you
run into difficulties with, say, the Monty Hall problem.  See \cite{GH02} for 
a discussion of this point in the synchronous setting.  I ignore this
issue here, since it is orthogonal to the issues that arise in the
Sleeping Beauty problem.} 

In any case, in asynchronous systems, the two probabilities may be
unequal for reasons beyond those that arise in the synchronous case.
This is perhaps best seen by considering a system 
where the agent might actually learn that it is Monday.  
The system $\R_2$ described Figure~\ref{fig:sb2} is one such system.     
Note that in $\R_2$, even if the HT approach is used, if
you discover it is Monday in run $r_1^*$ or $r_2^*$, 
then you do indeed ascribe probability $1/2$ to heads.
On the other hand, in $r_1$ and $r_2$,
where you do {\em not\/} discover it is Monday, you also ascribe
probability $1/2$ to heads when you are woken up, but conditional on it
being Monday, you consider the probability of heads to be $2/3$.
Thus, using the HT approach, $\R_2$ gives an example of a system where
the probability of heads given that it is Monday is different from the
probability of heads conditional on learning that it is Monday.

\commentout{
The real issue here is whether, in $r_1$, the probability
of heads conditional on it being Monday should be the same as the
probability that you ascribe to heads in $r_1^*$, where you actually
discover that it is Monday.  We often identify the probability of $V$
given $U$ with the probability that you would ascribe to $V$ if you
learn $U$ is true.  What I am suggesting here is that this
identification breaks down in the asynchronous case.  
This, of course, raises the question of what exactly
conditioning means in this context.  ``The probability of $V$ given
$U$'' is saying something like ``if it were the case that 
$U$, then the probability of $V$ would be \ldots''
This is not necessarily  the same
as ``if you were to learn that $U$, then the probability of $V$ would be
\ldots''%
}

Although $\R_2$ shows that Elga's argument for the $1/3$--$2/3$ answer
is suspect, it does not follow that $1/3$--$2/3$ is incorrect.  In the 
remainder of this section, I examine other considerations to see if they
shed light on what should be the appropriate answer.

\subsection{The Frequency Interpretation}

One standard interpretation of probability is in terms of frequency.  If
the probability of a coin landing heads is $1/2$, then if we repeatedly
toss the coin, it will land heads in roughly half the trials; it will
also land heads roughly half the time.  In the synchronous case, ``half
the trials'' and ``half the time'' are the same.  But now consider the
Sleeping Beauty problem.   What counts as a ``trial''?  If a ``trial''
is an experiment, then the coin clearly lands heads in half of the
trials.  But it is equally clear that the coin lands heads $1/3$
of the times that the agent is woken up.  Considering ``times'' and
``trials'' leads to different answers in asynchronous systems; in the
case of the Sleeping Beauty problem, these different answers are
precisely the natural $1/2$--$1/2$ and $1/3$--$2/3$ answers.
I return to this issue in the next subsection.


\subsection{Betting Games}

Another standard approach to determining subjective probability, which
goes back to Ramsey \citeyear{Ram} and De Finetti
\citeyear{DeFinetti31}, is in terms of betting behavior.  For example,
one way of determining the 
subjective probability that an agent ascribes to a coin toss landing
heads is to compare the odds at which he would accept a bet on heads to
one at which he would accept a bet on tails.  While this seems quite
straightforward, in the asynchronous case it is not.
This issue was considered in detail in the context of the
absented-minded driver paradox in \cite{GroveHalpern95}.  Much
the same comments hold here, so I just do a brief review.

Suppose that Sleeping Beauty 
is offered a \$1 bet on whether the coin landed heads or the coin
landed tails every time she is woken up.  If the bet pays off every time she
answers the question correctly, then clearly she should say ``tails''.
Her expected gain by always saying tails is \$1 (since, with probability
$1/2$, the coin will land tails and she will get \$1 both times she is
asked), while her expected gain by always saying heads is only $1/2$.  
Indeed, a risk-neutral agent should be willing to pay to take this bet.
Thus, even though she considers heads and tails equally likely and
ascribes probabilities using the HT approach, this
betting game would have her act as if she considered tails twice as likely as
heads:~she would be indifferent between saying ``heads'' and ``tails''
only if the payoff for heads was \$2, twice the payoff for tails.

In this betting game, the payoff occurs at every time step.  Now
consider a second
betting game, where the payoff is only once per trial
(so that if the coin lands tails, the agent get \$1 if she says tails
both times, and \$0.50 if she says tails only once).  
If the payoff
is per trial, then the agent should be indifferent being saying
``heads'' and ``tails''; the situation is analogous to the discussion in
the frequency interpretation.

There is yet a third alternative.
The agent could be offered a bet at only
one point in the information set.   If the coin lands heads, she must be
offered the bet at $(r_1,1)$.  If the coin lands heads, an adversary
must somehow choose if the bet will be offered at $(r_2,1)$ or
$(r_2,2)$.  The third betting game is perhaps more in keeping with the
second story 
told for $\R_1$, where the agent is not aware of time passing and must
assign a probability to heads and tails in the information set.  
It may seem that the first betting game, where the payoff occurs
at each step, is more appropriate to the Sleeping Beauty problem---after
all, the agent is woken up twice if the coin lands tails.  Of course, 
if the goal of the problem is to maximize the expected number of correct
answers (which is what this betting game amounts to), then there is no
question that ``tails'' is the right thing to say.  On the other hand,
if the goal is to get the right answer ``now'', whenever now is, perhaps
because this is the only time that the bet will be offered, then the
third game is more appropriate.  My main point here is that the
question of the right betting game, while noncontroversial in the
synchronous case, is less clear in the asynchronous case.

It is interesting to see how these issues play out in the context of
Hitchcock's \citeyear{hitchcock04} Dutch Book analysis of the Sleeping
Beauty problem.  As Hitchcock points out, there is a collection of bets
that form a Dutch book, 
which can be offered by a Bookie who knows no more than Sleeping Beauty
provided Sleeping Beauty ascribes probability
$1/2$ to heads when she wakes up:%
\footnote{The importance of taking the knowledge of the
Bookie into account, which is stressed by Hitchcock, is also one of the key
points in \cite{HT}.  Indeed, it is argued by Halpern and Tuttle that
probability does not make sense without taking the knowledge of the
adversary (the Bookie in this case) into account.} 
\begin{itemize}
\item Before the experiment starts, Sleeping Beauty is offered
a bet that pays off \$30 if the coin lands tails and 0 otherwise, and
costs \$15.  Since 
heads and tails are viewed as equally likely before the experiment
starts, this is a fair bet from her point of view.
\item Each time Sleeping Beauty is woken up, she is offered a bet
that pays off \$20 if the coin lands heads and 0 otherwise, and costs
\$10.  Again, if Sleeping Beauty views heads and tails as equally
likely when she is woken up, this bet is fair from her point of view.
\end{itemize}
Note that, if the coin lands heads, Sleeping Beauty is only waken up
once, so she loses \$15 on the first bet and has a net gain of \$10 on
the second bet, for an overall loss of \$5.  On the other hand, if the coin
lands heads, Sleeping Beauty has a net gain of \$15 on the first bet,
but the second bet is offered twice and she has a loss of \$10 each time
it is offered.  Thus, she again has a net loss of \$5.  

This Dutch Book argument is essentially dealing with bets that pay off
at each time step, since if the coin lands tails, Sleeping Beauty loses \$10
each time she is woken up.  By way of contrast, consider the
following sequence of bets:
\begin{itemize}
\item Before the experiment starts, Sleeping Beauty is offered
a bet that pays off \$30 if the coin lands heads and 0 otherwise, and
costs \$15.  
\item Each time Sleeping Beauty is woken up, she is offered a bet
that pays off \$30 if the coin lands tails and 0 otherwise, and costs
\$20, with the understanding that the bet pays off only once in each trial.
In particular, if the coin in fact lands tails, and Sleeping Beauty
takes the bet both times she is woken up, she gets the \$30 payoff only
once (and, of course, only has to pay \$20 for the bet once).  The
accounting is done at the end of the trial.
\end{itemize}
Note that the first bet is fair just as in the first Dutch Book, and the
second bet is fair to an agent who ascribes probability $2/3$ to tails
when woken up, even though the payoff only happens once if the coin
lands tails.  Moreover, although the second bet is somewhat nonstandard,
there is clearly no difficulty deciding when it applies and how to make
payoffs.   And, again, an agent who accepts all these bets will lose \$5
no matter what happens.

\subsection{Conditioning and the Reflection Principle}

To what extent is it the case
that the agent's probability over time can be viewed as changing via
conditioning?  It turns out that the answer to this question is closely
related to the question of when the Reflection Principle holds, and
gives further support to using the HT approach to ascribing
probabilities in the asynchronous case.

There is a trivial sense  in which updating is never done by
conditioning:  At the 
point $(r,m)$, agent $i$ puts probability on the space $\K_i(r,m)$; at
the point $(r,m+1)$, agent $i$ puts probability on the space
$\K_i(r,m+1)$.  These spaces are either disjoint or identical (since the
indistinguishability relation that determines $\K_i(r,m)$ and
$\K_i(r,m+1)$ is an equivalence relation).
Certainly, if they are disjoint, agent $i$ cannot be updating by
conditioning, since the conditional probability space is identical to
the original probability space.  And if the spaces are identical, it is
easy to see that the agent is not doing any updating at all; her
probabilities do not change.

To focus on the most significant issues, it is best to factor out time by
considering only  the probability ascribed to runs.  Technically, this amounts
to considering {\em run-based events}, that is sets $U$ of points with the
property that if $(r,m) \in U$, then $(r,m') \in U$ for all times $m'$.
In other words, $U$ contains all the points in a given run or none of them.
Intuitively, we can identify $U$ with the set of runs that have points
in $U$.  To avoid problems of how to assign probability in asynchronous
systems, I start by considering synchronous systems.  
Given a set $V$ of points, let $V^- =
\{(r,m):(r,m+1) \in V\}$; that is, $V^-$  consists of all the
points immediately preceding points in $V$.  The following result, whose
straightforward proof is left to the reader, shows that in synchronous
systems where the agents have perfect recall, the agents do essentially
update by conditioning.  The probability that the agent ascribes to an
event $U$ at time $m+1$ is obtained by conditioning the probability he
ascribes to $U$ at time $m$ on the set of points immediately preceding
those he considers possible at time $m+1$.

\thm\label{thm:conditioningpr}{\rm \cite{Hal31}}
Let $U$ be a run-based event
and let $\R$ 
be a synchronous system where the agents have perfect recall.  Then
$${\Pr}_{r,m+1,i}(U) =  {\Pr}_{r,m,i}(U \mid \K_i(r,m+1)^-).$$ 
\ethm

Theorem~\ref{thm:conditioningpr} does not hold without assuming perfect recall.   
For example, suppose that an agent tosses a fair coin and observes at
time 1 that the outcome is heads.
Then at time 2 he forgets the outcome (but remembers
that the coin was tossed, and knows the time).  Thus, at time 2, because
the outcome is forgotten, the
agent ascribes probability $1/2$ to each of heads and tails.  Clearly,
her time 2 probabilities are not the result of applying conditioning to
her time 1 probabilities.

A more interesting question is whether Theorem~\ref{thm:conditioningpr}
holds if we assume perfect recall and do not assume synchrony.  
Properly interpreted, it does, as I show below.  But, 
as stated, it does not, even with the HT approach to assigning
probabilities.  The problem is the
use of $\K_i(r,m+1)^-$ in the statement of the theorem.  In an
asynchronous system,  some of the points in $\K_i(r,m+1)^-$ may still be
in $\K_i(r,m+1)$, since the agent may not be aware of time passing.  
Intuitively, at time $(r,m)$, we want to condition on the set of points
in $\K_i(r,m)$ that are on runs that the agent considers possible at
$(r,m+1)$.  But this set is not necessarily $\K_i(r,m+1)^-$.  

Let $\K_i(r,m+1)^{(r,m)} = 
\{(r',k) \in \K_i(r,m): \exists m' ((r,m+1) \sim_i
(r',m'))\}$.  Note that $\K_i(r,m+1)^{(r,m)}$ consists precisely of those
points that agent considers possible at $(r,m)$ that are on runs that
the agent still  considers possible at $(r,m+1)$.  
In synchronous systems with perfect recall,
$\K_i(r,m+1)^{(r,m)} = \K_i(r,m+1)^-$ since, as observed above, if
$(r,m+1) \sim_i (r',m+1)$ then $(r,m) \sim_i (r',m)$.   In general, however,
the two sets are distinct. Using $\K_i(r,m+1)^{(r,m)}$ instead of
$\K_{r,m+1}^-$ gives an appropriate generalization of
Theorem~\ref{thm:conditioningpr}.

\thm\label{thm:conditioning}{\rm \cite{Hal31}} Let $U$ be a run-based event
and let $\R$ 
be a system where the agents have perfect recall.  Then, 
$${\Pr}_{r,m+1,i}^{{\it HT}}(U) =  {\Pr}_{r,m,i}^{{\it HT}}(U \mid \K_i(r,m+1)^{(r,m)}).$$ 
\ethm

Thus, in systems with perfect recall, using the HT approach to assigning
probabilities, updating proceeds by conditioning.
Note that since the theorem considers only run-based events, it holds
no matter how the probability among points on a run is distributed.  For
example, in the Sleeping Beauty problem, this result holds even if
$(r_2,1)$ and $(r_2,2)$ are not taken to be equally likely.

The analogue of Theorem~\ref{thm:conditioning} does not hold in general
for the Elga approach.
This can already be seen in the Sleeping Beauty problem.  Consider the
system of Figure~\ref{fig:sb1}.  At time 0 (in either $r_1$ or $r_2$),
the event heads (which consists of all the points in $r_1$) is ascribed
probability $1/2$.  At time 1, it is 
ascribed probability $1/3$.  Since $\K_{{\it SB}}(r_1,1)^{(r_1,0)} =
\{(r_1, 0), (r_2, 0)\}$, we have 
$$
1/3 = {\Pr}_{r_1,1,{\it SB}}^{{\it Elga}}({\it heads})  \ne 
{\Pr}_{r_1,0,{\it SB}}^{{\it Elga}}({\it heads}) \mid \K_{{\it
SB}}(r_1,1)^{(r_1,0)}) = 1/2.
$$ 
The last equality captures the intuition that if Sleeping Beauty gets no
additional information, then her probabilities should not change using
conditioning.

Van Fraassen's \citeyear{vF95a} {\em Reflection Principle\/} is a
coherence condition connecting an agent's future beliefs and his current
beliefs.  Note that what an agent believes in the future will depend in
part on what the agent learns.  The {\em Generalized Reflection
Principle\/} says that 
an agent's current belief about an event $U$ should lie in
the span of of the agent's possible beliefs about $U$ at some later time
$m$.    
That is, if $\Pr$ describes the agent's current beliefs, and $\Pr_1,
\ldots, \Pr_k$ describe the agent's possible beliefs at time $m$, then 
for each event $U$, $\Pr(U)$ should lie between $\min_j \Pr_j(U)$ and 
$\max_j \Pr_j(U)$.
Savage's \citeyear{Savage} {\em Sure-Thing  
Principle\/} is essentially a special case of the Generalized Reflection
Principle.  It says that if the
probability of $A$ is $\alpha$ no matter what is learned at time $m$,
then the probability of $A$ should be $\alpha$ right now.  This
certainly seems like a reasonable criterion.

Van Fraassen \citeyear{vF95a} in fact claims that if
an agent changes his opinion by conditioning on evidence, 
that is, if ${\Pr}_j = \Pr(\cdot  \mid E(j,m))$ for $j = 1, \ldots, k$,
then the Generalized Reflection Principle must hold.  The intuition is
that the pieces of evidence $E(1,m)$, \ldots, $E(k,m)$ must form a
partition of underlying space (in each state, exactly one piece of
evidence will be obtained), so that it becomes a straightforward
application of elementary probability theory to show that if $\alpha_j =
\Pr(E(j,t))$ for $j = 1, \ldots, k$, then  $\Pr = \alpha_1 {\Pr}_1 +
\cdots + \alpha_k {\Pr}_k$.

Van Fraassen was assuming that the agent has a fixed set $W$ of possible
worlds, and his probability on $W$ changed by
conditioning on new evidence.  Moreover, he was assuming that the
evidence was a subset of $W$.  
In the multiagent systems framework, the agent is not putting probability
on a fixed set of worlds.  Rather, at each time $k$, he puts probability
on the set of worlds (i.e., points) that he considers possible at time
$k$.  The agent's evidence is an information set---a set of points.
If we restrict attention to run-based events, we can instead focus on
the agent's probabilities on runs.  That is, we can take $W$ to be the
set of runs, and consider how the agent's probability on runs changes
over time.  Unfortunately, agent $i$'s evidence at a point $(r,m)$ is
not a set of runs, but a set of points, namely $\K_i(r,m)$.  We can
associate with $\K_i(r,m)$ the set of runs going through the points in
$\K_i(r,m)$, namely, in the notation of Section~\ref{sec:general},
$\R(\K_i(r,m))$

In the synchronous case, for each
time $m$, the possible information sets at time $m$ correspond to
the possible pieces of evidence that the agent has at time $m$.  These
information sets form a partition of the time-$m$ points, and induce a
partition on runs.  In this case, van Fraassen's argument is correct.
More precisely, if, for simplicity, ``now'' is taken to be time 0, and
we consider some future time $m > 0$, the possible pieces of evidence
that agent $i$ could get at time $m$ are all sets of the form
$\K_i(r,m)$, for $r \in \R$.  
With this translation of terms, it is an immediate consequence of van
Fraassen's observation and Theorem~\ref{thm:conditioningpr}
that the Generalized Reflection Principle
holds in synchronous systems with perfect recall.
But note that the assumption of perfect recall is critical here.
Consider an agent that tosses a coin and observes that it lands heads at
time 0.  Thus, at time 0, she assigns probability 1 to the event of that
coin toss landing heads.  But she knows that one year later she will
have forgotten the outcome of the coin toss, and will assign that event
probability $1/2$ (even though she will know the time).  
Clearly Reflection does not hold.


What about the asynchronous case?  Here it is not straightforward
to even formulate an appropriate analogue of the Reflection Principle.
The first question to consider is what pieces of evidence to consider at
time $m$.  While we can consider all the
information sets of form $\K_i(r,m)$, where $m$ is fixed and $r$ ranges
over the runs, these sets, as we observed earlier, contain points other
than time $m$ points.  While it is true that either $\K_i(r,m)$ is
identical to $\K_i(r',m)$ or disjoint from $\K_i(r',m)$, these sets do
{\em not\/} induce a partition on the runs.  It is quite possible that,
even though the set of points $\K_i(r,m)$ and $\K_i(r',m)$ are disjoint,
there may be a run $r''$ and times $m_1$ and $m_2$ such that $(r'',m_1)
\in \K_i(r,m)$ and $(r'',m_2) \in \K_i(r',m)$.  
For example, in Figure~\ref{pchdyn.fig3}, if the runs from left to right
are $r_1$--$r_5$, then $\K_{\mathit{SB}}(r_5,1) = \{r_1, \ldots, r_5\}$
and $\K_{\mathit{SB}}(r_1,1) = \{r_1, r_2, r_3\}$.
However,
under the assumption of perfect recall, it can be shown that for any two
information sets $\K_i(r_1,m)$ and $\K_i(r_2,m)$, either (a)
$\R(\K_i(r_1,m)) \subseteq \R(\K_i(r_2,m))$, (b) $\R(\K_i(r_2,m)) \subseteq
\R(\K_i(r_1,m))$, or (c) $\R(\K_i(r_1,m)) \inter \R(\K_i(r_2,m)) =
\emptyset$. From 
this it follows that there exist a collection $\R'$ of runs such that
the sets  $\R(\K_i(r',m))$ for $r' \in \R'$  are disjoint and
the union of $\R(\K_i(r',m))$ taken over the runs $r' \in \R'$ consists of
all runs in $\R$.  Then the same argument as in the 
synchronous case gives the following result.

\thm\label{pro:reflectionOK} If $\R$ is a (synchronous or asynchronous)
system with perfect recall and 
$\K_i(r_1,m)$, \ldots, $\K_i(r_k,m)$ are the distinct information sets of the
form $\K_i(r',m)$ for $r' \in \R(\K_i(r,0)$, then there 
exist $\alpha_1, \ldots, \alpha_k$ such that 
$${\Pr}_{i}(\cdot \mid \R(\K_i(r,0))) = \sum_{j=1}^k
\alpha_j {\Pr}_i(\cdot \mid \R(\K_i(r_j,m))).$$
\ethm

The following corollary is immediate from
Theorem~\ref{pro:reflectionOK}, given the definition of
$\Pr_{(i,r,m)}^{{\it HT}}$. 
\cor\label{cor:reflectionOK} 
If $\R$ is a (synchronous or asynchronous) system with
perfect recall and $\K_i(r_1,m)$, \ldots, $\K_i(r_k,m)$ are the distinct
information sets of the 
form $\K_i(r',m)$ for $r' \in \R(\K_i(r,0)$, then there 
exist $\alpha_1, \ldots, \alpha_k$ such that for all $\R' \subseteq \R$,
$${\Pr}_{(i,r,0)}^{{\it HT}} (\K_i(r,0)(\R')) = \sum_{j=1}^k
\alpha_j {\Pr}^{{\it HT}}_{(i,r_j,m)}(\K_i(r_j,m))(\R')).$$
\ecor

Corollary~\ref{cor:reflectionOK} makes precise the sense in which the
Reflection Principle holds for the HT approach.  Although the notation
$\K_i(r,m)(\R')$ that converts sets of runs to sets of points makes the
statement somewhat ugly, it plays an important role in emphasizing what
I take to be an important distinction, that has largely been ignored.
An agent assigns probability to points, not runs.  At both time 0 and
time $m$ we can consider the probability that the agent assigns to the
points on the runs in $\R'$, but the agent is actually assigning
probability to quite different (although related) events at time 0 and
time $m$.  It is important to note that I am not claiming here that
$\alpha_j = \Pr(\R(\K_i(r_j,m))$ in Theorem~\ref{pro:reflectionOK}.
While this holds in the synchronous case, it does not hold in general.
The reason we cannot expect this to hold in general is that, in the
synchronous case, the sets $\R(\K_i(r_j,m))$ are disjoint, so 
$\sum_{j=1}^n \Pr(\R(\K_i(r_j,m)) = 1$.  This is not in general true in
the asynchronous case.  I return to this issue shortly.

The obvious analogue to Corollary~\ref{cor:reflectionOK} does not hold
for the Elga approach.  Indeed, the same example that shows
conditioning fails in the Sleeping Beauty problem shows that the
Reflection  Principle does not hold. Indeed, this example shows that the
sure-thing principle fails too.  Using the Elga approach, the
probability of heads (i.e., the probability of the points on the run
where the coin lands heads) changes from $1/2$ to $1/3$ between time 0
and time 1, no matter what.

Arntzenius \citeyear{Art03} gives a number of other examples where he
claims the 
Reflection Principle does not hold.  In all of these examples, 
the agent either has imperfect recall or the system is
asynchronous and the Elga approach is being used to ascribe
probabilities.   Thus, his observation may not seem surprising, given
the previous analysis.  However, in one case,
according to my definition, Reflection in fact does not fail.  This is
due to the fact that I interpret Reflection in a slightly different way
from Arntzenius.  Since this example is of independent interest, I now
consider it more carefully.  


The example, credited by Arntzenius to John Collins, is the
following:  A prisoner has in his cell two clocks, both of which
run perfectly accurately.  However, clock $A$ initially reads 6 PM and
clock $B$ initially reads 7 PM.  The prisoner knows that exactly one of
the clocks is accurate; he believes that with probability $1/2$ the
accurate clock is clock $A$ and with probability $1/2$ it is clock $B$.
The prisoner also knows that a fair coin has been tossed to determine if the
lights go out at midnight; if it lands heads, they do, and if it lands
tails, they stay on.  Since the coin is fair, the prisoner initially
places probability $1/2$ on it landing heads.

There are four runs in the system corresponding to this problem, each
of which has probability $1/4$:
\begin{itemize}
\item $r_1$, where $A$ is the accurate clock and the coin landed heads;
\item $r_2$, where $A$ is the accurate clock and the coin landed tails;
\item $r_3$, where $B$ is the accurate clock and the coin landed heads;
\item $r_4$, where $B$ is the accurate clock and the coin landed tails.
\end{itemize}
We can assume that the environment state encodes the true time and the
outcome of the coin toss, while
the prisoner's state encodes the clock readings and whether the light is
off or on.  Thus, a typical global state might have the form
$((\mbox{11:30}, H), (\mbox{11:30}, \mbox{12:30}, 1))$.  In this global state, the true time is
\mbox{11:30} and the coin landed heads, clock $A$ reads \mbox{11:30}
(and is correct), 
clock $B$ reads \mbox{12:30}, and the light is on (denoted by the component
1 in the tuple).  Thus, this is the global state at the point
$(r_1, \mbox{11:30})$.  The other points in the same information set as
$(r_1,\mbox{11:30})$ are $(r_2, \mbox{11:30})$ and $(r_4, \mbox{12:30})$.  Call this
information set $I_1$.  At all the three points in $I_1$, 
the prisoner's local state is $(\mbox{11:30}, \mbox{12:30}, 1)$.
For future reference, note that the only other information set that
includes time \mbox{11:30} points is $I_2 = \{(r_1,\mbox{10:30}), (r_2, \mbox{10:30}), (r_3,
\mbox{11:30}), (r_4,\mbox{11:30})\}$.  At all the points in $I_2$, the
pair of clocks read 
$(\mbox{10:30}, \mbox{11:30})$ and the light is on.  

It is easy to check that every information set has at most one point per
run.  It follows from Lemma~\ref{lem:identical} that at every point,
the HT approach and the Elga approach agree.  Thus, no matter which
approach is used, Reflection in the sense of
Corollary~\ref{cor:reflectionOK} must hold.  Observe that 
the prisoner's degree of belief that the coin landed heads in
information set $I_1$ is $2/3$, while in $I_2$ it is $1/2$.  
Thus, the prisoner's initial probability of heads ($1/2$) is a convex
combination of his possible probabilities of heads at 11:30, but the
combination has coefficients 0 and 1. Taking the coefficients to be 0
and 1 might seem a little strange.  After all, why should we prefer 
$I_2$ so strongly?  But I claim that the 
``strangeness'' here is a result of carrying over inappropriate
intuitions from the synchronous case.
In the synchronous case, the coefficients
reflect the probability of the information sets.  This makes sense in
the synchronous case, because the information sets correspond to
possible pieces of evidence that can be obtained at time $m$, and the
sum of these probabilities of the pieces of evidence is 1.  However, in
the asynchronous case, we cannot relate the coefficient to probabilities
of obtaining evidence.  Indeed, the ``evidence'' in the case of
information set $I_2$ is that the clock readings are (10:30, 11:30) and
the light is on.  This is evidence that the prisoner initially knows
that he will certainly obtain at some point (although not necessarily at
11:30).  Indeed, it falls out of the analysis of
Theorem~\ref{pro:reflectionOK} that it   
does not make sense to relate the coefficients in the asynchronous case
to the probabilities of obtaining the evidence. 

Arntzenius points out another anomaly in this example.
Taking $P_t$ to denote the prisoner's probability at (real) time
$t$, Arntzenius observes that $${\Pr}_{7:00} (\mbox{clock $B$ is correct}
\mid {\Pr}_{11:30} (\mbox{clock $B$ is correct}) = 1/3) = 0.$$
For $\Pr_{11:30} (\mbox{clock $B$ is correct}) = 1/3)$ holds only
in runs $r_1$ and $r_2$, since at the points $(r_1,11:30)$ and
$(r_2,11:30)$, the prisoner's probability that $B$ is correct is $1/3$,
while at the points $(r_3,11:30)$ and $(r_4,11:30)$, the prisoner's
probability that $B$ is correct is $1/2$.  On the other hand, $B$ is
not correct in runs $r_1$ and $r_2$, so the conditional probability is
0.

Arntzenius suggests that this is a problem, since the prisoner does not
trust his later beliefs.  I would argue that the prisoner should trust
all his later beliefs that he is aware of.  The trouble is, the prisoner
has no idea when he has the belief $\Pr_{11:30}(\mbox{clock $B$ is
correct}) = 1/3$, since he has no idea when it is 11:30.  
(Essentially the same point is made by Schervish, Seidenfeld, and Kadane
\citeyear{SSK04}.)  Of course, in a synchronous system, an agent does know
when 11:30 is, so beliefs of the form $\Pr_{11:30}(U)$ are ones he
should trust.

\commentout{
It is certainly true that the expected probability of heads in the two
situations that the prisoner considers possible when it is actually
11:30 is not $1/2$.  However, despite appearances, this is not a
violation of the Reflection Principle.  The prisoner has no way of
knowing when it is actually 11:30.  The only evidence that the prisoner
has is the clock readings.  Reflection requires conditioning on the
evidence.  It is easy to check that, at 6 PM, the prisoner ascribes
probability 1 to reaching a situation where the readings are
(10:30, 11:30) and the lights are on.  In this situation, as we have
seen, the probability of heads is $1/2$, so there is no violation of
Reflection with this evidence.  The prisoner also ascribes probability 
$3/4$ of reaching a reading of (11:30, \mbox{12:30}) where it is light (this
will happen if either $A$ is the correct clock or if the coin landed
tails) and a probability of $1/4$ of reaching a reading of (11:30, 12:30)
where it is dark.  In the former case, he ascribes probability $1/3$ to
heads, as we have seen; in the latter case, he ascribes probability 1 to
heads.  Thus, the expected probability of heads according to
the prisoner when he has a reading of (11:30, 12:30) is $1/2$, exactly as
it should be according to the Reflection Principle. 
}

Note that if we modify the problem very slightly so that 
(a) clock $A$ gives the true time, (b) the lights will be turned off when the
jailer's clock reads midnight, and (c) one of $A$ and $B$ gives the
jailer's time, but the prisoner does not know which and ascribes each
probability $1/2$, then we get a
synchronous system which is identical to Arntzenius's in all essential
details.  However, now Reflection is completely unproblematic.
At 11:30, if the light is still on, the prisoner ascribes
probability $1/3$ to heads; if the light is off, the prisoner
ascribes probability 1 to heads.  Initially, the prisoner ascribes
probability $3/4$ to the light being on at 11:30 and probability $1/4$
to the light being off. Sure enough $1/2 = 3/4 \times 1/3 + 1/4 \times
1$.  

This example emphasizes how strongly our intuitions are based on
the synchronous case, and how our intuitions can lead us astray in the
presence of asynchrony.  The prisoner has perfect recall in this system,
so the only issue here is synchrony vs.~asynchrony.

\section{Conclusion}\label{sec:conc}

In this paper, I have tried to take a close look at the problem of
updating in the presence of asynchrony and imperfect recall.
Let me summarize what I take to be the main points of this paper:
\begin{itemize}
\item It is important to have a good formal model that
incorporates uncertainty, imperfect recall, and asynchrony in which
probabilistic arguments can be examined.  While the model I have
presented here is certainly not the only one that can be used, it does
have a number of attractive features.
As I have shown elsewhere \cite{Hal15}, it can also be used to deal with
other problems involved with imperfect recall raised by Piccione and
Rubinstein \citeyear{PR97}. 
\item Whereas there seems to be only one reasonable approach to
assigning (and hence updating) probabilities in the synchronous case, there
are at least two such approaches in the asynchronous case.  Both
approaches can be supported using a frequency interpretation and a betting
interpretation.  However, only the HT approach supports the Reflection
Principle in general.  In particular,
the two approaches lead to the two different answers
in the Sleeping Beauty problem. 
\item We cannot necessarily identify the probability conditional on $U$
with what the probability would be upon learning $U$.  This
identification is being made in Elga's argument; the structure $\R_2$
shows that they may be distinct.
\end{itemize}

One fact that seems obvious in light of all this discussion is that our 
intuitions regarding how to do updating in asynchronous systems are
rather poor.  
This is clearly a topic that deserves further
investigation. 

\section*{Acknowledgments}  Thanks to Moshe Vardi for pointing out 
Elga's paper, 
to Teddy Seidenfeld for pointing out Arntzenius's paper,
to Moshe, Teddy, Oliver Board, and Sergiu Hart for stimulating 
discussions on the topic, and to 
Oliver, Moshe, Adam Elga, Alan H{\'a}jek, James Joyce, Kevin O'Neill,
and two anonymous KR reviewers 
for a number 
of useful comments on an earlier draft of the paper. 

\bibliographystyle{chicago}
\bibliography{z,joe}
\end{document}